\documentclass[sigconf]{acmart}

\usepackage{multicol}
\usepackage{multirow}
\usepackage{color}
\usepackage{balance}
\usepackage{float}
\usepackage{stfloats}

\AtBeginDocument{%
  }

\copyrightyear{2024} 
\acmYear{2024} 
\setcopyright{rightsretained} 
\acmConference[MM '24]{Proceedings of the 32nd ACM International Conference
 on Multimedia}{October 28-November 1, 2024}{Melbourne, VIC, Australia}
 \acmBooktitle{Proceedings of the 32nd ACM International Conference on
 Multimedia (MM '24), October 28-November 1, 2024, Melbourne, VIC, Australia}
\acmDOI{10.1145/3664647.3681400}
 \acmISBN{979-8-4007-0686-8/24/10}

\begin{document}

\title{Towards Flexible Evaluation for Generative \\ Visual Question Answering}

\author{Huishan Ji, Qingyi Si, Zheng Lin\textsuperscript{\textasteriskcentered}}
\email{{jihuishan, siqingyi, linzheng}@iie.ac.cn}
\orcid{0009-0002-7815-9508, 0000-0001-8433-0215, 0000-0002-8432-1658}
\affiliation{%
  \institution{Institute of Information Engineering, \\ Chinese Academy of Sciences}
  \institution{School of Cyber Security, \\ University of Chinese Academy of Sciences}
  \city{Beijing}
  \country{China}
}

\author{Weiping Wang}
\email{wangweiping@iie.ac.cn}
\orcid{0000-0002-8618-4992}
\affiliation{%
  \institution{Institute of Information Engineering, \\ Chinese Academy of Science}
  \city{Beijing}
  \country{China}
}

\thanks{\textsuperscript{\textasteriskcentered}Corresponding author: Zheng Lin, linzheng@iie.ac.cn}

\renewcommand{\shortauthors}{Huishan et al.}

\begin{abstract}
Throughout rapid development of multimodal large language models, a crucial ingredient is a fair and accurate evaluation of their multimodal comprehension abilities. Although Visual Question Answering (VQA) could serve as a developed test field, limitations of VQA evaluation, like the inflexible pattern of Exact Match, have hindered MLLMs from demonstrating their real capability and discourage rich responses. 
Therefore, this paper proposes the use of semantics-based evaluators for assessing unconstrained open-ended responses on VQA datasets. As characteristics of VQA have made such evaluation significantly different than the traditional Semantic Textual Similarity (STS) task, to systematically analyze the behaviour and compare the performance of various evaluators including LLM-based ones, we proposes three key properties, i.e., Alignment, Consistency and Generalization, and a corresponding dataset Assessing VQA Evaluators (AVE) to facilitate analysis. 
In addition, this paper proposes a Semantically Flexible VQA Evaluator (SFVE) with meticulous design based on the unique features of VQA evaluation.
Experimental results verify the feasibility of model-based VQA evaluation and effectiveness of the proposed evaluator that surpasses existing semantic evaluators by a large margin. The proposed training scheme generalizes to both the BERT-like encoders and decoder-only LLM. Relaed codes and data available at \href{https://github.com/jihuishan/flexible_evaluation_for_vqa_mm24}{Our Repository}.
\end{abstract}



\received{13 April 2024}
\received[accepted]{21 July 2024}

\maketitle

\section{Introduction} \label{sec_introduction}
Visual Question Answering (VQA) evaluates the multimodal comprehension abilities 
by posing questions about given images and comparing the model's responses with annotated answers\cite{daquar,coco-qa,vqav2,okvqa,gqa,gurari2018vizwiz,fvqa}. 
However, current VQA evaluation metrics have made it tough for evaluating the rich responses of Multimodal Large Language Models (MLLMs).

Most VQA datasets comply with a triplet format and each sample consists of a question, image and annotation. Annotations are often a single word or phrase \cite{daquar,coco-qa, gqa, fvqa, kbvqa} or a set of ten candidate answers \cite{schwenk2022aokvqa, okvqa, gurari2018vizwiz, vqav2}. Current evaluation metrics, Exact Match \cite{daquar} (for samples without candidate answers) and VQA Score \cite{vqa} (for samples with ten candidate answers), both require the responses to be identical in morphology with the annotation to be considered correct. Variations in tense, singular or plural forms and synonyms are not allowed, let alone sentence-style responses from MLLMs. 

\begin{figure}[t]
\begin{center}
   \includegraphics[width=0.7\linewidth]{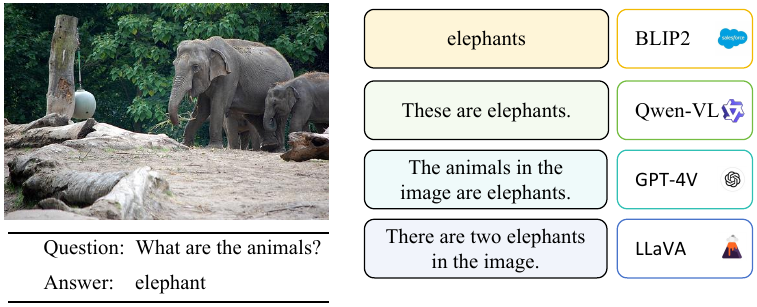}
\end{center}
   \caption{Responses from four MLLMs on a simple visual question. The responses are different in length, styles and complexity, which can all be considered correct but none of them exactly matches the annotated answer.\vspace{-0.7cm}}
\label{mllm_responses}
\end{figure}

Traditional vison-language models treat VQA as a classification problem \cite{noh2016image, wu2017image, yang2016stacked,coco-qa, LXMERT}, where answers collected from the training set are used to establish pre-defined classes, and the possible responses are constrained to these classes. Thus the problem of evaluating multifarious responses does not exist. However, MLLMs treat VQA as a generative problem \cite{li2023blip, instructblip, Qwen-VL, llava1.5liu2023improved} and generates assorted responses. Meanwhile, the growing trend that the MLLM community prefers zero-shot test, has made it even tougher for models to generate responses that are identical to the ground-truth answers. As shown in Figure \ref{mllm_responses}, semantically equivalent but morphologically distinct responses are not accepted.



Although it is possible to force the model to output a single word with a harsh prompt, such remediation may potentially damage the performance and make it unfair for different MLLMs, especially for those with poor instruction-following ability and those that tend to response with long sentences. 
As MLLMs inherit the in-context learning capability of LLMs, it is feasible to introduce in-context examples to force short responses. However, since different MLLMs contain different in-context learning capability, such practice interferes a fair evaluation of MLLMs' multimodal comprehension performance.
There thus is an urgent need for a metric that aligns well with human judgment and accommodates various response types while ensuring consistent evaluation despite variations in response morphology.


To compare different evaluators, traditional Semantic Textual Similarity (STS) task 
measures the difference between predicted scores and human annotation results from a single aspect of semantic relevance. However, both intuition and our experiments (refer to Section \ref{background} and \ref{main_experiments}) suggest that there is a significant difference between the evaluation of correctness in VQA responses and the traditional assessment of media-text relevance in STS. 

Therefore, to systematically evaluate the performance of an evaluator, with the unique characteristics in the task of VQA response evaluation taken into consideration, we propose three quantitative key properties, i.e., \textbf{Alignment}, \textbf{Consistency} and \textbf{Generalization}. Alignment stands for the overall correspondence of predicted scores with human annotation. Consistency measures how well an evaluator accommodates semantically equivalent responses of different morphology and length. Generalization indicates the variance of performance on different sources of data. Further, to facilitate comprehensive analysis of the performance and behaviour of evaluators, we provide a human-annotated dataset Assessing VQA Evaluators (AVE) that grades the correctness of model responses towards ground-truth labels on VQA datasets. AVE is further augmented by ChatGPT and WordNet \cite{miller1990introductionwordnet} to increase the diversity.

As pilot experiment shows (refer to Section \ref{main_experiments}), formulaic metrics (BLEU \cite{papineni2002bleu}, ROUGH \cite{lin2004rouge}, METEOR \cite{banerjee2005meteor}) and model-based metrics \cite{gao2021simcse, li2023angle, reimers2019sentence, bge_m3} perform poorly on the VQA response evaluation task. Therefore, we propose a novel evaluator that is trained with meticulously designed pretraining tasks. The tasks are designed for improving the embedding representation of VQA text, which utilizes contrastive learning to guide the evaluator to capture the fine-grained difference within a text pair and ignore the noise in morphology and length.
Experiments demonstrate that the proposed pretraining tasks significantly improve the performance of our evaluator on the AVE dataset, making the evaluator's prediction aligns much better with human judgement.


The contribution of this paper can be concluded as follows:

\begin{itemize}

    \item This paper addresses the dilemma, where rich responses of MLLMs hinder fair evaluation under current metric, by proposing semantic-similarity-based evaluation that applies to various VQA responses.
    \item This paper proposes three quantitative key properties in VQA response evaluation based on its characteristics, and a high-quality human-annotated dataset, AVE, for assessing different evaluators comprehensively. 
    In addition, we evaluate the performance of various types of existing semantic similarity evaluators on the proposed AVE dataset.
    \item Experimental results demonstrate the feasibility of applying model-based methods to the flexible evaluation of VQA responses as well as the effectiveness of our proposed evaluator. Our evaluator significantly surpasses existing methods, including ChatGPT and the SOTA embedding model Voyage-lite-02-Instruct \footnote{As of the time of submission, Voyage-lite-02-Instruct achieves the best performance on the task of Semantic Textual Similarity (STS).} by a large margin. Our training scheme generalizes to both encoder-only and decoder-only models.

\end{itemize}

\section{Related Work}

\subsection{Visual Question Answering} 

As the answer space of most open-ended VQA datasets is limited and the same answer applies for multiple questions (the most common 2,000 answers in the training set of VQA v2\cite{vqav2} is able to cover about 94\% questions in its validation set), early methods \cite{noh2016image, wu2017image, yang2016stacked,coco-qa, LXMERT} treat VQA as a classification task, which adopt answers in the training set as class labels, unable to predict unseen classes. Generative methods on VQA \cite{instructblip, llavaliu2024visual, Qwen-VL,li2023blip, llava1.5liu2023improved} treat VQA as a generation task and facilitates much more variant responses. 

\subsection{Semantic Textual Similarity} 
\label{sts_related_work}
Current semantic evaluation tasks include Semantic Textual Similarity (STS) \cite{agirre2012semeval, agirre2013sem} that assesses to what extent the two sentences are related, Paraphrase Identification \cite{tan2018multiway, yin2016abcnn} that decides whether two texts express the same meaning, and Natural Language Inference \cite{williams2017broadmnli, conneau2017supervisedsnli} that determines the logical relationship between texts. The essence of these tasks lies in quantifying the degree of semantic equivalence between sentences, which is a fundamental challenge due to the complexity and variability of natural language. Methods in STS include formulaic methods like BLEU \cite{papineni2002bleu} and model-based ones \cite{gao2021simcse, li2023angle, 2024instruct, reimers2019sentence, zhang2019bertscore}. The former mainly relies on n-gram or other statistic features between the candidate and reference to calculate the overlap and import penalty for noise. The latter utilizes models as encoders to extract the information and compare between the candidate and reference. Early model-based evaluator \cite{zhang2019bertscore} compares the similarity of each pair each time, which is computation-consuming. Later works \cite{gao2021simcse, li2023angle, 2024instruct, reimers2019sentence} first generate embeddings separately for the candidate and reference, then simply calculate the cosine similarity between them as the similarity score. Due to the style of STS, either formulaic or model-based methods pay more attention to the overall similarity and are less capable of detecting fine-grained semantic difference, as shown in our experiments. 

\subsection{Multimodal Comprehension Evaluation of MLLMs}
As a developed realm and valuable resource of high-quality data, VQA has been applied in the evaluation of MLLMs. Several works \cite{fu2023mme, ying2024mmt, schwenk2022aokvqa} propose multiple-choice datasets to make evaluation simple and straight. For example, MME \cite{fu2023mme} proposes a smart quantitative analysis of MLLMs with manually designed instruction-answer pairs that strictly limit responses to be \textit{yes} or \textit{no}. Therefore, all MLLMs are evaluated relatively fairly. However, although MME is insightful and effective, such detour avoids the problem of evaluating open-ended response directly. It ignores previous huge amount of VQA data and costs additional human annotation, limiting the scale of the dataset and making it tough for expanding. MM-vet \cite{yu2023mm-vet}, LVLM-eHub \cite{shao2023tiny} and ConvBench \cite{liu2024convbench} classify VQA into the integration of multiple key abilities, and manually annotates corresponding VQA samples of their required abilities. Then, they use ChatGPT for evaluation (which we show to be less capable in evaluating the correctness of open-ended responses, refer to Section \ref{main_experiments}). The classification of VQA abilities is insightful and aids to the probing of specific abilities of MLLMs. Yet MM-vet requires high-quality annotation to identify the VQA abilities each question requires and are thus limited to a small amount too.

\section{Semantic Evaluation of VQA}
With the rapid development of MLLMs, current metrics in VQA response evaluation are too stubborn to assess the rich generation and hinder evaluating MLLMs' performance with existing VQA datasets. Meanwhile, as mentioned in Section \ref{sts_related_work}, current semantic evaluation models and tasks are inconsistent with the goal of flexible VQA response evaluation.



Therefore, this paper proposes the task of semantic evaluation of VQA, aiming at introducing flexible similarity-based soft evaluation with continuous scores into the assessment of VQA responses, contrary to inflexible metrics like Exact Match or VQA Score that require identical morphology of responses towards ground-truth labels. Such flexible evaluation enables to assess the rich generation from MLLMs and thus enables to use existing VQA datasets for probing MLLMs' multimodal comprehension ability.



\subsection{Characteristics of VQA Evaluation} \label{background}
The proposed task of semantic VQA response evaluation shares significant difference with existing semantic evaluation tasks like STS and contains its own characteristics.

\paragraph{Discrimination Granularity} As mentioned in Section \ref{sec_introduction}, traditional semantic evaluation tasks typically focus on the overall meaning in texts, rather than capturing the fine-grained detailed difference. However, the core of semantic VQA evaluation is comparing the response with annotated answer under the same question\footnote{Considering the polysemy and ambiguity of words and phrases, the question text is indispensable for evaluating the semantic correctness.}, where both texts share large overlap in meaning as the questions are same, demanding fine-grained semantic discrimination. 

\paragraph{Text Length} As VQA answers are generally much shorter than texts in STS \footnote{About 97.9\% answers in VQA v2 \cite{vqav2}, 97.7\% in OKVQA \cite{okvqa}, 99.9\% in GQA \cite{gqa} are shorter than three words. The average length of text in STS-12\cite{agirre2012semeval} is 12.5, which is much longer.}, n-gram based formulaic metrics like BLEU \cite{papineni2002bleu} will be more easily affected by the context in response. Model-based metrics are also vulnerable to such length shift, as their training data barely cover similar pattern.

\paragraph{Distribution Shift} The texts in STS datasets \cite{agirre2012semeval,agirre2013sem,agirre2014semeval} come from general domains, like news and social media, while different VQA datasets comply to different sub-tasks, like knowledge \cite{okvqa} or reasoning \cite{gqa}. Such distribution shift causes inconsistent evaluation on responses from different VQA datasets. 

\subsection{Three Key Properties in VQA Evaluation} \label{key_properties}
To systematically evaluate the performance of a VQA evaluator, we propose three quantitative key properties, i.e., \textbf{Alignment}, \textbf{Consistency} and \textbf{Generalization}.

\paragraph{Alignment} Alignment assesses the overall performance of similarity scores predicted by evaluators with that of human annotation, in the metric of Spearman's Rank Correlation following similar setting in previous works\cite{gao2021simcse, li2023angle, agirre2012semeval, agirre2013sem}.
\paragraph{Consistency} A smart evaluator shall catch the key information in responses and ignore the noise text, e.g., the response of \textit{elephants} shall be scored equally with \textit{Theses are elephants} under the question of \textit{What are the animals?}. Therefore, Consistency measures how close the different responses sharing the same meaning are scored.
\paragraph{Generalization} Considering various VQA datasets focus on various sub-tasks and come from various sources, Generalization depicts how well an evaluator is able to handle text from different domains. Refer to Section \ref{task_definition} for quantitative definitions.

\subsection{A Dataset Assessing VQA Evaluators} 
\label{dataset_Parts}

To comprehensively compare and analyze the behaviour of different evaluators on VQA responses, taking the proposed three key properties into consideration, we propose a dataset Assessing VQA Evaluators (AVE). 
By collecting multiple MLLMs' responses on multiple datasets, the proposed dataset simulates a real scene of applying evaluators to evaluate the quality of various VQA responses. In order to compare the evaluators' scoring results with human judgement, we provide human annotation of the semantic correctness of responses towards ground-truth answers. The construction process of AVE is shown as follows:


\paragraph{Response Collection} First, we collect responses of five models, LLaVA \cite{llavaliu2024visual}, BLIP2 \cite{li2023blip}, mPLUG-Owl \cite{ye2023mplug}, OFA-large \cite{wang2022ofa}, Qwen-VL \cite{Qwen-VL} on the validation set of four datasets, OKVQA\cite{okvqa}, A-OKVQA\cite{schwenk2022aokvqa}, VQA v2\cite{vqav2} and GQA\cite{gqa} (balanced testdev set). 

\paragraph{Sampling Results} Second, we sample in the responses while controlling the sampling amount of each dataset to be the same. In addition, samples that are answered correctly, i.e., the response is identical with the ground-truth answer, are excluded.

\paragraph{Human Annotation} Third, three annotators are asked to measure the semantic similarity\footnote{We considered multiple aspects of measuring the correctness of a response towards the ground-truth answer, yet at last we come to the single aspect of semantic similarity for annotation. Refer to Appendix for more explanation.} of each sampled response towards the ground-truth label and annotate an integral similarity score from 0 to 10, under certain rules (refer to Appendix for more details). Then the scores are averaged over the three annotators.

\paragraph{Description Generation} Fourth, in order to simulate MLLM responses with sentences instead of words or phrases, we select responses that are shorter than three words for augmentation \footnote{For longer responses, we manually condense it.}. The augmentation is conducted in two ways. The first way comes from using ChatGPT (refer to Appendix for prompts) to convert each pair of question and response into three descriptions and asking ChatGPT to select two descriptions that are closest to the original question-answer pair as augmented responses. For example, the question of \textit{What are the animals? } and the response of \textit{elephants} are fed into ChatGPT, and it generates descriptions like \textit{The animals are elephants.} The second way comes from applying manually designed answer templates (refer to Appendix) to increase the diversity of descriptions rather than fully relying on ChatGPT. Now each sample contains three descriptions, i.e., the two ChatGPT-augmented descriptions and a manual-template-fitted one.

\paragraph{Synonym Generation} Fifth, we use WordNet \cite{miller1990introductionwordnet} to locate a synonym for each answer. For cases where multiple synonyms exist, we choose the most common synonym by countering frequencies of words in Brown Corpus \cite{francis1979brown}. Meanwhile, we ask ChatGPT to introduce a shift in morphology to simulate cases that the outputs are merely different in tenses or singular or plural forms. The augmented answer is deemed to contain the same meaning with small disturbance on the style. 

\paragraph{Manual Filter} At last, to ensure high-quality of the dataset, the three annotators also conduct manual filter (refer to Appendix for rules) to eliminate ambiguous samples, especially those generated from the fifth stage.

\begin{figure}[t]
\begin{center}
   \includegraphics[width=1\linewidth]{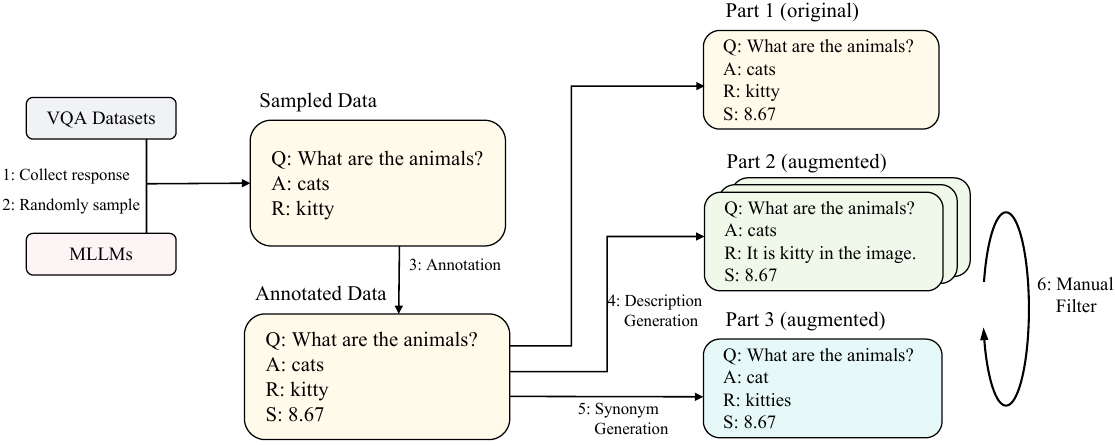}
\end{center}
   \caption{The construction procedure of AVE. After randomly sampled from the outputs of models, each sample is manually annotated with a score and automatically augmented by generated descriptions and a variation on the answer word while remaining almost the same correctness as a VQA response. Part 1 to 3 denote different augmentation methods. \vspace{-0.4cm}} 
\label{AVE}
\end{figure}

Generally speaking, the whole AVE dataset consists of three parts generated from above: Part 1 contains original answers and original responses (without augmentation). Part 2 contains original answers and the generated three descriptions of responses. Part 3 contains tense-shifted answers and original responses. The whole procedure is depicted in Figure \ref{AVE}. 

Meanwhile, the AVE dataset can also be clustered by the involved four datasets that each sample belongs to, OKVQA, A-OKVQA, VQA v2 and GQA, merging the Part 1 to Part 3 together and classifies by the sources of data only. Refer to Section \ref{task_definition} for how they are used. 

The total sample amount of the final AVE dataset is 3,592, with each sample containing four types of augmentation results, as described above. The dataset is then split into a validation set and a test set with the ratio of 3:7. 
The distribution of annotated scores is shown in Appendix, which is relatively smooth. 
To evaluate the inter-annotator agreement, following previous works \cite{grusky2018newsroom, fabbri2021summeval, bhandari2020realsumm}, we apply Krippendorff's alpha \cite{krippendorff2011computing} and obtain a result of 0.713.

\subsection{The Proposed Evaluation Indicators} \label{task_definition}
In AVE dataset, a sample consists of a question $q_i$, a ground-truth label $a_i$, a source dataset label $d_i$, a response $r_i$ and a human-annotated score $s_i$. The task of VQA response evaluation can be defined as: given a question-answer pair, an evaluator $f(q_i, a_i, r_i)$ is expected to predict the annotated similarity score $s_i$ with the output $o_i$.

\begin{equation}
o_i = cos(f(q_i, r_i), f(q_i, a_i))
\end{equation}

\begin{equation}
score_f = Spearman(O, S)
\end{equation}

where $cos$ stands for cosine similarity, $score_f$ is the performance score of the evaluator $f$, with $O$ and $S$ indicating the lists of all predicted scores and annotated scores respectively. Note that the metrics used for evaluating evaluators is Spearman's rank coefficient of correlation (Spearman).

The key properties of Alignment, Consistency and Generalization introduced in Section \ref{key_properties} are computed as follows:
\paragraph{Alignment} We use the average result of an evaluator on all parts of the proposed AVE dataset as alignment:

\begin{equation}
    Alignment = \frac{1}{N_{Parts}}\sum^{N_{Parts}}_{i=1} score_{f_i}
\end{equation}

where $N_{Parts}$ is the number of Parts in the AVE dataset and $N_{sets}$ is the number of involved VQA datasets in AVE, according to different type of division. The $score_{f_i}$ is the spearman score of evaluator $f$ on the $i$ th part of AVE. 

\paragraph{Consistency} Consistency measures how close the responses of the same meaning with different morphology are evaluated. We regard the variance of the same sample among the three parts of AVE as consistency:

\begin{equation}
    Consistency = \log{(1/( \frac{1}{N_{samples}}\sum^{N_{samples}}_{j=1}var(o_{j_1}, o_{j_2}, o_{j_3})))}
\end{equation}

where $N_{samples}$ is the amount of samples and var denotes calculating the variance. Then, $o_{j_1}$, $o_{j_2}$, $o_{j_3}$ are the predicted scores of the evaluator on Part 1, 2, 3 for the same sample $j$, respectively.

\paragraph{Generalization} Generalization measures the difference of performance on various datasets, and we define it as the variance of the performance on each involved VQA dataset:

\begin{equation}
\begin{split}
\text{Generalization} = \log 1/\ var( align_{OKVQA}, \\
align_{A-OKVQA}, align_{VQAv2}, align_{GQA} )
\end{split}
\end{equation}

where $align_{dataset}$ is the mean Alignment score on the AVE data belonging to the corresponding VQA dataset.

\section{Semantically Flexible VQA Evaluator}

With the three key properties of an ideal evaluator taken into consideration, we propose a novel evaluator based on meticulously designed pretraining tasks.


\subsection{Pretraining Tasks}
To guide the model to be sensitive to the key information between answer and response, this paper introduces several pretraining tasks to enhance the embedding. Data for augmentation come from a random sampling of VQA data in the training sets of OKVQA\cite{okvqa}, A-OKVQA\cite{schwenk2022aokvqa}, TDIUC\cite{tdiuc}, VG-QA\cite{vg}, GQA\cite{gqa} and VQA v2 \cite{vqav2}, which end to a total amount of 105, 311 samples. All augmented samples are mixed for training.

\paragraph{NLI data} In previous works \cite{gao2021simcse, li2023angle}, models perform well with the natural language inference datasets SNLI \cite{conneau2017supervisedsnli} and MNLI \cite{williams2017broadmnli}, where each sample includes a premise, an entailment and a contradiction.
In addition, these NLI datasets are all manually constructed, ensuring the high quality of their data, and the premise shares limited overlap with entailment compared with the sentence pairs in back-translation datasets. 
Therefore, to ensure the fundamental discriminating ability of models to capture overall meaning of sentences, we adopt NLI data and regard the premise-entailment pairs as postive pairs and premise-contradiction pairs as negative pairs.

\paragraph{Candidate answers} To make the best of available VQA datasets, for datasets with ten candidate answers, OKVQA, A-OKVQA and VQA v2, we consider candidate answers as correct answers as well. Then, for each sample, the most common candidate answer and a less common one are used to form a positive pair, with a random answer sampled from the answer space as negative. 

\paragraph{Synonym and Antonym} In VQA response evaluation, semantically similar answers shall receive similar scores. We replace the answer with a synonym by WordNet \cite{miller1990introductionwordnet}, and if the antonym of an answer exists, we then pair up the answer and antonym as a negative pair, else we pair up the answer and a randomly sampled answer from the answer space as a negative pair. In addition, 
we use ChatGPT to produce synonyms as well, as ChatGPT is able to capture contextual information in the question and thus generates more accurate synonyms.

\paragraph{Generated descriptions} To simulate the output of MLLMs, for each sample, we provide ChatGPT with its question and answer to generate three descriptions with small disturbance of the same meaning. Then, we construct positive samples by pairing up the original answer and each generated description. For negative samples, we replace the answer in the generated description with a randomly sampled answer. The goal is to pull the embedding representation of a natural language description close to its simple form of a single answer, so that responses with different length but carrying similar meanings will receive similar scores. In addition, the negative pair is constructed by replacing the key answer word in the description, therefore guiding the model to be sensitive to the key words and to ignore the noise.

\subsection{Model Framework}

\begin{figure*}[t]
\begin{center}
   \includegraphics[width=0.7\linewidth]{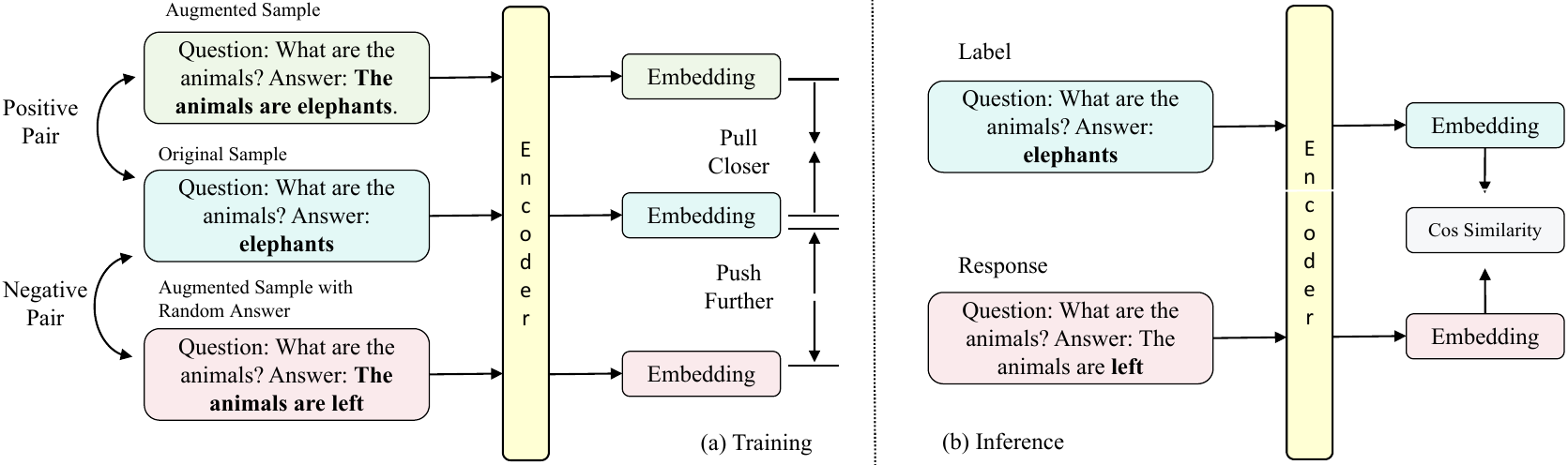}
\end{center}
   \caption{Framework of contrastive learning in the proposed Semantically Flexible VQA Evaluator (SFVE). The original sample is augmented into two variations and form a positive pair and a negative pair. The example in the figure shows the procedure of the pretraining task \textit{Generated descriptions}. In the positive pair, the semantics of the sentence is considered same as the original, while in the negative pair, as the answer word is replaced with a random answer, the sentence contains unmatched meaning with the original.\vspace{-0.4cm}} 
\label{model_framework}
\end{figure*}

Following previous works \cite{chen2020simple, gao2021simcse, conneau2018senteval, li2023angle} on the STS task \cite{agirre2012semeval, agirre2013sem, agirre2014semeval, agirre2015semeval, agirre2016semeval, cer2017semevalstsb}, we use cosine similarity for distance calculation between embeddings\cite{gao2021simcse, li2023angle}.
As shown in Figure \ref{model_framework}, we adopt the simple contrastive learning framework \cite{chen2020simple} and contrastive learning with in-batch hard negatives \cite{gao2021simcse}. 

The backbone encoder in this paper is RoBERTa \cite{liu2019roberta}. In order to gain better generalization and comprehension ability, we apply the decoder-only LLM LLAMA2 \cite{touvron2023llama2} with the prompt\cite{li2023angle} of \textit{Summarize the text \{text\} in a single word:}. Then, the hidden states of the first generated new token is considered as the embedding vector.

The contrastive learning with in-batch hard negatives loss \cite{gao2021simcse} is defined as follows:

\begin{equation}
    loss_{ibn} = -\log{\frac{e^{sim({h_i,h_i^+})/\tau}}{\sum^N_{j=1}(e^{sim({h_i,h_j^+})/\tau}+e^{sim({h_i,h_i^-})/\tau}})}
\end{equation}

where $h_i$ is the embedding representation of sample $i$, $h_i^+$ and $h_i^-$ respectively denote the representation of the positive sample and in-batch hard negative sample of sample $i$.

\section{Experiments}

\subsection{Implementation Details}
Experiments in this paper is based on \textit{transformer} package\cite{wolf-etal-2020-transformers} on Pytorch. We use AdamW \cite{adamw} optimizer, and the hyper-parameters of AdamW, betas, eps and weight-decay are set to 0.9, 0.999, 1e-8 and 0.01. We use a cosine scheduler and the batch size and peak learning rate for encoders are 128, 1e-5 for RoBERTa-base \cite{liu2019roberta}, VisualBERT \cite{li2019visualbert} and LXMERT \cite{LXMERT}, 32, 6e-6 for RoBERTa-large and 8, 4e-6 for LLAMA2 \cite{touvron2023llama2}. 

\subsection{Baselines} \label{baselines}
To assess to what extent existing models are competent for the VQA response evaluation, this paper collects four types of common methods for semantic similarity evaluation and refer to them as: formulaic, PLM, LLM, and API.
\begin{itemize}
    \item Formulaic methods contain BLEU \cite{papineni2002bleu}, ROUGE \cite{lin2004rouge} and METEOR \cite{banerjee2005meteor}. These methods base on n-grams for assessing the overlap. As VQA answers are usually short, we also report 2-gram results for BLEU and ROUGE. 
    \item PLM refers to the Pretrained Language Models, which are generally small in sizes and typically in BERT-like encoder-only structures. 
    SBERT (Sentence BERT) \cite{reimers2019sentence} embeds texts into vectors with BERT and apply cosine similarity to measure the distance as textual similarity. 
    SIMCSE \cite{gao2021simcse} provides both unsupervised and supervised methods, and this paper selects the supervised and better-performing one trained on NLI datasets for comparison. 
    BGE \cite{bge_m3} follows a multi-task learning scheme that collects and pretrains on multifarious datasets for better generalization.
    AnglE \cite{li2023angle} aims to mitigate the gradient saturation issue encountered when using cosine distance by projecting vectors onto the complex plane and introducing an angular loss. 
    \item LLM refers to large language models. This paper selects four of the well-performing LLMs, Baichuan2 \cite{baichuan2023baichuan2}, Qwen \cite{qwen}, LLAMA-2 \cite{touvron2023llama2} and Mistral \cite{Mistral7b}.
    \item API refers to the remote usage of models by API online, including ChatGPT and Text-embedding-v3-large from OpenAI, and Voyage-lite-02-instruct from Voyage AI. Refer to Appendix for prompts. The latter two are embedding models that produce text embedding of given text, which are then used to calculate similarity score by cosine distance.
\end{itemize}

\vspace{-0.4cm}

\subsection{Main Experiments} \label{main_experiments}

\begin{table*}[] 
\centering
\scalebox{0.7}{
\begin{tabular}{ c | c | c c c c | c | c | c} 
    \toprule
    \multirow{2}*{Types} & \multirow{2}*{Methods} & \multicolumn{4}{c|}{Alignment $\uparrow$}  & \multirow{2}*{Consistency $\uparrow$}  & \multirow{2}*{Generalization $\uparrow$} & \multirow{2}*{STS Avg. $\uparrow$} \\
    
    &  & Part 1 & Part 2 & Part 3 & Avg. &  &  & \\ \hline
    \multirow{5}*{Formulaic} & BLEU-2 &  -1.7 & -2.6 & 3.2 & -0.4 & 4.35 & 11.30 & 50.1 \\
     & BLEU-4 & -2.9 & -3.5 & 2.0 & -1.5 & 4.23 & 11.25 & 47.6 \\
     & ROUGE-2 &  -2.8 & -4.5 & 1.2 & -2.0 & 5.79 & 11.82 & 53.9 \\
     & ROUGE-L &  4.9 & -0.1 & 3.3 & 2.7 & 6.73 & 10.78 & 48.3 \\ 
     & METEOR & 12.4 & 4.4 & 15.3 & 10.7 & 7.25 & 9.34 & 53.4 \\\hline
    \multirow{5}*{PLM} & RoBERTa-large (w/o CL) & 11.9 & 0.7 & 23.4 & 12.0 & 7.91 & 10.36 & 27.9\\
     & SBERT & 47.7 & 44.3 & 40.6 & 44.2 & 8.79 & 8.55 & 76.8\\
     & SIMCSE & 44.9 & 44.7 & 41.7 & 43.7 & 9.37 & 8.30 & 83.8 \\
     & BGE & 42.3 & 36.5 & 41.0 & 39.9 & 8.93 & 8.25 & 84.9 \\ 
     & AnglE & 43.4 & 38.2 & 40.2 & 40.6 & 9.01 & 7.78 & 86.4 \\ \hline
    \multirow{4}*{LLM} & Baichuan2-7b & 28.1 & 27.8 & 31.8 & 29.2 & 5.30 & 8.86 & 64.6 \\
     & Qwen-7b & 25.9 & 26.3 & 24.2 & 25.5 & 9.07 & 10.10 & 68.3\\
     & LLaMA2-7b & 32.7 & 27.9 & 34.6 & 31.7 & 7.45 & 8.53 & 61.9 \\ 
     & Mistral-7b & 16.8 & 14.5 & 20.7 & 17.3 & 4.61 & 8.49 & 72.1 \\ \hline

    \multirow{3}*{API} & ChatGPT & 21.2 & 15.2 & 24.6 & 20.3 & 5.21 & 8.35 & 73.7 \\ 
     & Text-embedding-v3-large & 32.5 & 28.6 & 36.3 & 32.5 & 9.40 & 8.00 & 82.3 \\
     & Voyage-lite-02-instruct & 29.1 & 28.9 & 29.3 & 29.1 & 11.81 & 6.78 & 86.3 \\ \hline
    \multirow{3}*{SFVE (ours)} & SFVE-base & 58.4 & 57.1 & 53.7 & 56.4 & 9.12 & 8.34 & 81.2 \\
     & SFVE-large & 58.1 & 57.5 & 56.0 & 57.2 & 9.53 & 8.67 & 82.0 \\
     & SFVE-LLAMA2-7b & 60.2 & 57.0 & 57.2 & 58.1 & 9.46 & 8.87 & 77.9 \\

    \bottomrule

    \end{tabular}
    }
\caption{The comparison of performance on our proposed AVE dataset. The STS Avg. denotes the average scores over STS 2012 to STS 2016 \cite{agirre2012semeval, agirre2013sem, agirre2014semeval, agirre2015semeval, agirre2016semeval}, SICK-R \cite{marelli2014sickr} STS-B \cite{cer2017semevalstsb}, providing a reference of methods' general discriminating ability. RoBERTa-large \cite{liu2019roberta} (w/o CL) refers to the original pretrained checkpoint without contrastive learning. SFVE-base, SFVE-large and SFVE-LLAMA2-7b are RoBERTa-base, RoBERTa-large and LLAMA2-7b trained by contrastive learning on our proposed pretraining tasks. The specific model checkpoints in experiments are as follows: SBERT\cite{reimers2019sentence}: SRoBERTa-NLI-large, BGE\cite{bge_m3}:BAAI-bge-large-en, SIMCSE\cite{gao2021simcse}: RoBERTa-NLI-large, AnglE \cite{li2023angle}: RoBERTa-large.\vspace{-0.8cm}} 

\label{main_experiments_table}
\end{table*}

\begin{table*}[] 
\centering
\scalebox{0.8}{
\begin{tabular}{ c | c c c c | c | c} 
    \toprule
    \multirow{2}*{Settings} & \multicolumn{4}{c|}{Alignment $\uparrow$}  & \multirow{2}*{Consistency $\uparrow$}  & \multirow{2}*{Generalization $\uparrow$} \\
     
    & Part 1 & Part 2 & Part 3 & Avg. &  & \\ \hline

    All tasks & 58.4 & 57.1 & 53.7 & 56.4 & 9.12 & 8.34 \\ \hline
    w/o \textit{NLI data} & 53.3 & 52.3 & 50.9 &52.2 & 8.71 & 8.50 \\
    w/o \textit{Candidate answers} & 57.3 & 56.5 & 53.0 & 55.6 & 9.33 & 8.54\\
    w/o \textit{Synonym and Antonym} & 42.1 & 40.3 & 38.8 & 40.4 & 9.19 & 8.41\\
    w/o \textit{generated descriptions} & 56.9 & 47.0 & 52.3 & 52.1 & 8.07 & 8.93\\ \hline
    w/o All tasks & 12.5 & 3.1 & 20.0 & 11.8 & 10.58 & 9.90 \\ \hline
    \textit{NLI data} only & 44.3 & 42.0 & 33.1 & 39.8 & 9.46 & 10.00  \\
    \textit{Candidate answers} only & 37.4 & 29.8 & 39.6 & 35.6 & 8.10 & 8.11 \\
    \textit{Synonym and Antonym} only & 53.8 & 43.7 & 50.6 & 49.4 & 7.89 & 9.08 \\
    \textit{Generated descriptions} only & 42.8 & 49.1 & 42.1 & 44.6 & 8.17 & 7.67 \\
    \bottomrule
    \end{tabular}
    }

\caption{Ablation experiments of designed pretraining tasks on RoBERTa-base. The row of All tasks represents the best performance of RoBERTa-base with all pretraining tasks, and the row of w/o All tasks contains results from testing on the RoBERTa-base checkpoint without further training. w/o represents without the corresponding pretraining task, contrary to the setting in lower part of the table where the model is trained only on a single task each time.\vspace{-0.7cm}} 
\label{ablation}
\end{table*}

Table \ref{main_experiments_table} exhibits the performance comparison of various evaluators. The results are on the test set of AVE, with specific scores on each part of the dataset, as described in Section \ref{dataset_Parts}. To promote a comprehensive assessment of existing methods, this paper compares the performance with four common types of methods for semantic evaluation, as introduced in Section \ref{baselines}. The last row of Types contains our results from training with the proposed pretraining tasks on the corresponding model. Then, the column of Alignment contains the separate results on each of the three parts in AVE datasets and their average.

\subsubsection{Performance of Formulaic Methods}
Formulaic methods, i.e., BLEU, ROUGE and METEOR perform poorly in Alignment scores, and some of them drop below 0, indicating 
adverse scores to the human annotation. Such phenomenon is expected, as the n-gram matching strategy of BLEU and ROUGE is unable to handle the synonyms or variations in tenses and singular or plural forms. For METEOR, however, it applies port stem \cite{banerjee2005meteor} and synonym matching to preprocess the 1-gram in both the candidate and reference, restoring words to stems and thus performs better. 

In addition, it is interesting to notice that although the Alignment of BLEU and ROUGE are much lower that that of METEOR, their Generalization scores are much higher. There are two reasons to this anomaly. First, BLEU and ROUGE fail to handle the task well and their prediction can be considered random, thus the sources of data do not affect the results, just like RoBERTa-large (w/o CL). Second, these n-gram evaluators do not involve semantics, therefore the sources of data that causes word distribution shift matter less. 

\subsubsection{Performance of PLMs}
The BERT-like models pretrained for textual similarity prediction, i.e., SBERT, SIMCSE, BGE, AnglE (the latter four models), show much better performance than RoBERTa-large (w/o CL) and formulaic methods, indicating the basic textual similarity tasks are helpful to the VQA response evaluation task, but they fail to align well with human judgement, compared to SFVE results under the same structure of BERT-large. In addition, the performance on Part 1 and 3 of PLM models are similar, and the major gap lies in the capability of processing long responses. 

\subsubsection{Performance of LLMs}
For LLMs (refer to Appendix for the detailed prompt) including ChatGPT, they fail to gain satisfactory results on AVE. Naturally, LLM performs better than RoBERTa w/o CL, and Generalization scores are slightly higher than PLMs, which we attribute to the better generalization ability of LLMs. Although LLMs obtain acceptable results on STS tasks, just like the formulaic methods, they encounter significant performance drop on the VQA response evaluation. Such phenomenon verifies the significant difference between the evaluation of STS and VQA responses and the necessity in the task of VQA evaluation.

The performance of embedding models (the latter two models) on STS is higher than LLMs but the VQA response evaluation performance is still low. The reason to their incompetence on AVE, as we speculate, is that these models focus more on retrieving and capturing the general meaning of given texts than discovering the fine-grained difference between given pair of texts. In addition, such focus of capturing the general meaning has also empowered them with the ability to ignore noise in morphology and text length, thus gaining high scores of Consistency despite the low Alignment scores.

\subsubsection{Performance of SFVE}
The section of SFVE (ours) in the table presents our results on AVE. The pretraining tasks effectively improve the Alignment scores of all three models and bring moderately better Consistency and Generalization performance. From the prospective of model sizes, the 125M Roberta-base demonstrates similar capability with the 355M Roberta-large with merely a gap of 0.8\%. The same applies for the 7b LLAMA2, which surpasses RoBERTa-large by 0.9\%. Giant increase in model sizes brings minor improvement in scores. 
We believe the reason is that the similarity measure, either in STS or AVE, is relatively simple for models to comprehend and implement, where a simple structure with limited parameters achieves excellent performance with proper training. 

Therefore, for practical usage of evaluators during the training of generative VQA models, considering the significantly larger computation cost in LLAMA2-7b than RoBERTa, we recommend utilizing SFVE-RoBERTa base or large for a rough validation of model performance each certain steps or epoch, and use SFVE-LLAMA2 for more accurate evaluation near the best steps or epochs.


\subsection{Ablation Experiments}

To analyze the influence of each pretraining task, Table \ref{ablation} provides ablation results by removing a pretraining task each time and by training on a single task alone. From the table it is clear that all pretraining tasks contribute to the final performance more or less. 

The most important task is \textit{Synonym and Antonym}, which causes a drop of 16.0\% in Alignment scores on average and damages Consistency as well. In addition, when trained only on such data, the model performs the best. We believe the importance of training on \textit{Synonym and Antonym} task lies in aligning the representation of synonyms and increasing the difference towards antonyms and other answers.

The second influencing pretraining task is \textit{Generated descriptions}, without which the model can not directly learn to align the representation between semantically similar texts with different length. Yet the removal of it does not substantially damage the results on other parts than Part 2, which consists of long responses.

Meanwhile, the removal of \textit{NLI data} matters almost the same as \textit{Generated descriptions}. As mentioned before, NLI data focuses more on the coarse-grained meaning between text pairs while AVE requires a finer semantic discrimination. However, for a model that barely handles the task (shown in the row of \textit{w/o All tasks}), we believe the easier data in NLI aid to fertilizing the basic capability in semantic evaluation. Yet the NLI data alone is insufficient, as shown in the row of \textit{NLI data only}.

\begin{figure*}[ht]
\begin{center}
   \includegraphics[width=0.95\linewidth]{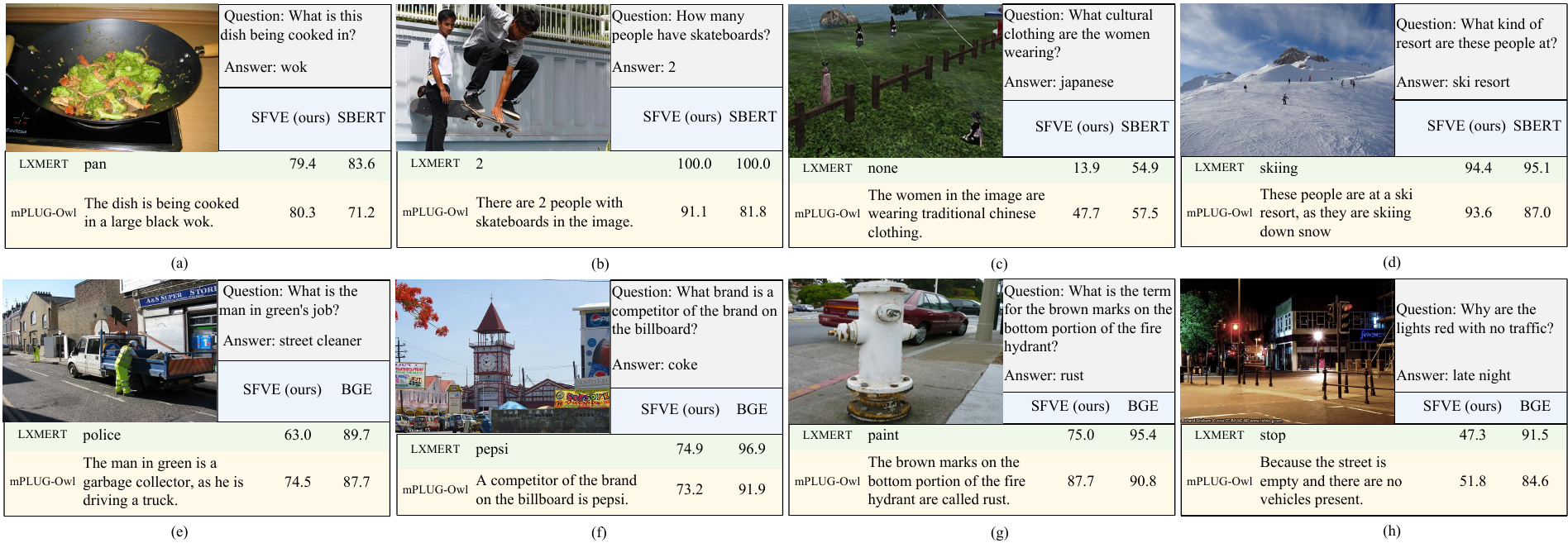}
\end{center}
   \caption{Cases for analysis. The samples come from the open-ended part of A-OKVQA \cite{schwenk2022aokvqa} validation set. The first row comes from results of SFVE-large and SBERT, and the second comes from SFVE-large and BGE.} 
\label{figure_case_study}
\end{figure*}
\balance
\subsection{Practical Application}
    
To demonstrate the practical values of our proposed evaluator in flexible VQA response evaluation, we collect responses of multiple MLLMs and compare the results with different evaluators by overall scores and case study.

\begin{table}[] 
\centering
\scalebox{0.7}{
\begin{tabular}{ c | c c c c} 
    \toprule
    \multirow{2}*{Model} & \multicolumn{4}{c}{Evaluation Metric} \\
     & VQA Score & BGE & SBERT & SFVE (ours) \\ \hline
    LXMERT $^\dag$ \cite{LXMERT} & 19.5 & 83.3 & 75.6 & 43.6  \\
    LXMERT  & 37.3 & 94.9 & 83.6 & 67.9 \\
    VisualBERT \cite{li2019visualbert} & 37.6 & 94.8 & 83.4 & 66.3 \\ \hline
    LLaVA-7b \cite{llavaliu2024visual} & 3.6 & 89.1 & 82.5 & 72.3 \\
    BLIP2-opt-2.7b \cite{li2023blip} & 15.5 & 94.1 & 83.1 & 70.2 \\
    InstructBLIP-Vicuna \cite{instructblip} & 21.4 & 94.8 & 86.4 & 74.3 \\
    mPLUG-Owl \cite{ye2023mplug} & 0.0 & 91.0 & 82.7 & 69.1 \\
    OFA-large \cite{wang2022ofa} & 39.5 & 95.3 & 86.5 & 78.0 \\
    Qwen-VL-chat \cite{Qwen-VL} & 54.9 & 96.1 & 89.7 & 83.5 \\
     
    \bottomrule
    \end{tabular}
    }

    \caption{Practical application of utilizing our proposed evaluator for assessing the responses from MLLMs. The VQA dataset for response generation is the open-ended validation set of A-OKVQA\cite{schwenk2022aokvqa}. Models in the upper part of the table are smaller than 0.5B. 
    VisualBERT and LXMERT are fine-tuned on VQA v2\cite{vqav2}. LXMERT $^\dag$ means the LXMERT that is not sufficiently trained, which ends training at the half of the first epoch to provide comparison. SFVE (ours) uses the RoBERTa-large evaluator trained with our proposed pretraining tasks. Refer to Appendix for the calculation of VQA Score. Note that the scores are for comparison within an evaluator itself, and it is meaningless to compare scores across evaluators, as evaluators are not aligned. \vspace{-0.6cm} }
\label{table_prac_application}
\end{table}

As shown in Table \ref{table_prac_application}, VQA score is clearly incompetent for assessing assorted responses from MLLMs. Since all responses from mPLUG-Owl are sentences, VQA Score even comes to 0. In the comparison of LXMERT and mPLUG-Owl, both BGE and SBERT indicate LXMERT generates better responses than mPLUG-Owl. However, taking the case study in Figure \ref{figure_case_study} into consideration, we verify that existing well-performing methods, BGE and SBERT, fail to perform consistent evaluation and bias towards short responses while penalizing longer ones. For example, in (a) of Figure \ref{figure_case_study}, LXMERT response \textit{pan} receives a much higher score than the mPLUG-Owl response which is a descriptive sentence containing the correct answer. In (d), the descriptive text and the single word response receive similar scores under our SFVE, but SBERT considers the short answer of LXMERT is much better than the descriptive sentence of mPLUG-Owl. Similar phenomena exist in BGE as well. As in (e), mPLUG-Owl response describing \textit{garbage collector} is a better response than LXMERT output \textit{police}, yet the latter receives even higher scores. 

In addition, not only does the length impede a fair evaluation, but the incompetence in fine-grained semantics discrimination also causes absurd results. Like in (c) and (g), where LXMERT answers are less correct but they receive competitive or even higher scores than reasonable responses from BGE and SBERT. Such error, we speculate, is caused by focusing more on the overall meaning of text, as the questions are the same within a pair, while neglecting fine-grained difference.

Due to the phenomena above, it is clear the superficial superiority of LXMERT over mPLUG-Owl is merely a mistake by incompetent VQA evaluators, which also demonstrates the importance of fairness and consistency in VQA evaluation. Therefore, we consider the proposed pretraining tasks and SFVE effective, not only on our proposed AVE dataset, but also in practical application where previous methods fail to perform fair and insightful evaluation. 

\section{Conclusion}
This paper proposes a practical task of utilizing semantic correctness to evaluate unconstrained open-ended VQA responses, facilitating the assessment of MLLMs' multimodal comprehension abilities by VQA data.
We propose three key properties for assessing VQA evaluators, i.e., Alignment, Consistency and Generalization.
In addition, this paper proposes a new dataset assessing VQA evaluators (AVE) to comprehensively analyze multiple aspects of evaluators. Based on contrastive learning with meticulously designed pretraining tasks, this paper provides a Semantically Flexible VQA Evaluator (SFVE) that performs significantly better than existing evaluators on VQA evaluation and the training scheme generalizes to both the encoder-only and decoder-only models.

\clearpage

\bibliographystyle{ACM-Reference-Format}
\bibliography{sample-base}

\appendix

\section{Traditional VQA Evaluation Metrics}
Traditional VQA evaluation metrics contain Exact Match \cite{daquar} and VQA Score \cite{vqa}. They apply for different settings in VQA datasets. For datasets where each sample contains only one correct answer, like DAQUAR \cite{daquar}, TDIUC \cite{tdiuc}, GQA \cite{gqa}, the Exact Match is used. If each sample contains ten candidate answers, like VQA v2 \cite{vqav2}, OKVQA \cite{okvqa}, VizWiz \cite{gurari2018vizwiz}, VQA Score is commonly used.

\paragraph{Exact Match} Exact Match calculates by judging whether the response is identical to the annotated ground-truth answer, and if matches, the score is 1, otherwise 0.

\paragraph{VQA Score} VQA Score evaluates how many times the response appear in the ten candidate answers, and is computed as follows:
$$
accuracy = min(\frac{\#\ correct\ hits}{3},1)
$$
As there are ten candidate answers, \# correct hits represents numbers of matched answers, which means as long as there are three or more candidates are the same with the predicted answer, the answer will be considered fully correct, and gets a score of 1. 

\section{Prompt for Decoder Models}
The following is the similarity calculation prompt provided to the LLMs and ChatGPT in the experiments:

\textit{Sentence similarity evaluation here refers to the task of measuring the semantic similarity score between two sentences. For example, "what a good day" and "how nice the weather is" are almost the same, your output shall be \{"score":0.91\}. Now please evaluate the similarity score between the following two sentences: sentence1: sample["sentence1"]. sentence2:  sample["sentence2"]. The score shall range continuously from 0-1. DO NOT output anything else but the .json-style dictionary, like \{"score":x\}, where x is your predicted score.}

The sample["sentence1"] and sample["sentence2"] indicate a pair of texts for similarity calculation. The question and answer are concatenated with the prompt "Question: \{question\} Answer: \{answer\}" before similarity calculation, importing contextual information, just the same as other models in experiments.

ChatGPT prompt for converting each question-answer pair into a description:

Concatenate the question with the answer and form assertions. For example, Question:What kind of dog is in the photo?  Answer:golden retriever.  Assertion: The dog in the photo is a golden retriever. Infer for the following: Question: \{question\} Answer: \{answer\}. Please think of three different forms of naturally-sounded assertions for this question-answer pair with small disturbance but do not output them. Choose the two assertions that are closest in meaning to the original question-answer for output. Output shall be in .json style so that I can directly save them in a .txt and open by json. Do not output anything else including explanation, reasoning or instructions.

\section{Annotation Score Distribution}

\begin{figure}[t]
\begin{center}
   \includegraphics[width=0.7\linewidth]{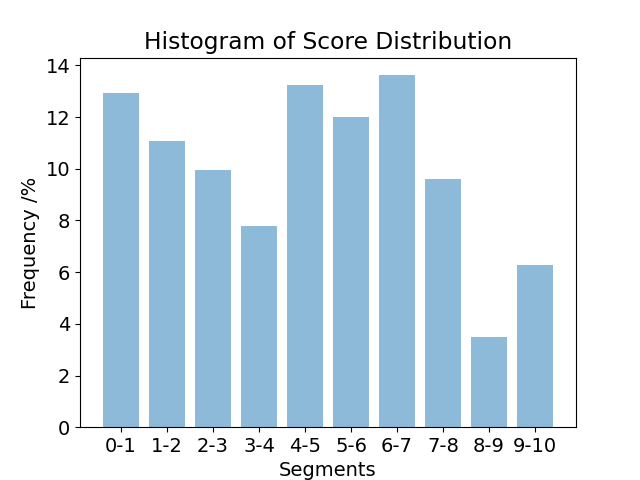}
\end{center}
   \caption{The annotated scores distribution of AVE.} 
\label{AVE_score_distribution}
\end{figure}

The score distribution of the dataset Assessing VQA Evaluators (AVE) is shown in Figure \ref{AVE_score_distribution}. The annotation covers all scores.

\section{Selection of Annotating Aspect}

For the evaluation of the quality of responses, we considered adopting multiple aspects for analyzing. The text summary task \cite{grusky2018newsroom, bhandari2020realsumm, fabbri2021summeval} adopts four aspects, i.e., \textit{relevance}, \textit{consistency}, \textit{fluency} and \textit{coherence} for analyzing the generated summary based on the reference text. As VQA responses are generally short, the \textit{fluency} and \textit{coherence} that measure the fluency of the text is less necessary. The aspect of \textit{consistency} measures whether the generated summary contains hallucination that generates untrue information. In the scene of VQA response evaluation, such measure corresponds to the overlap in semantics from the response towards the ground-truth answer, which is similar to \textit{relevance} that measures how well the generated text captures the key points. Therefore, we decide to use semantic similarity as the only score of annotation.

\section{Annotation Rules}

\subsection{Scores Annotation}
All of the following in this section is the annotation rules provided to the three annotators during the annotation of scores. 

Scoring Format: Discrete integer scoring from 0 to 10 (e.g., 0, 1, 2, 3).

Scoring Rules:

Note that it is not about judging whether the response to the question is correct, but whether the response and the standard answer are semantically the same under the question (For example, even if the standard answer is obviously unreasonable, as long as the answer and response are semantically similar, a high score shall be given).

Semantically similar (the answer is fairly correct in meaning) but different in specific form (for example, the meaning expressed is similar but different in word choice, tense, number), score 6-10 based on the degree of semantic similarity.
Examples:

(1) Question: What is the last letter on the license plate?
Standard Answer: letter j
Response: j
Scoring (this is a reasonable range, just mark a specific score when actually annotating): 9-10
Reason (no need to mark, this is to help understand the rules): Under this question, the response and the standard answer are semantically the same, only the form is different. Similarly, synonyms should also be scored highly.

(2) Question: The young man above the swimming pool is wearing what?
Standard Answer: swimsuit
Response: trunks
Scoring: 7-8
Reason: The question asks what is being worn, and swimsuit (swimwear) includes trunks (swim shorts), so the answer is quite correct. This kind of inclusive or included relationship should be scored based on the semantic similarity of the two words. In addition, trunks as swim shorts is a less common meaning and requires more attention to the different meanings of words, not entirely based on experience.

(3) Question: How many trees are there?
Standard Answer: 3
Response: three
Scoring: 10
Reason: The meanings are exactly the same.

Semantically dissimilar, the answer is incorrect, but the answer is a possible answer for that type of question, score 1-5 based on the degree of semantic similarity.
Examples:

(1) Question: What color do you think the trousers the boy is wearing have?
Standard Answer: white
Response: blue
Scoring: 2-3
Reason: The question asks about color, and although the answer blue is different from the standard answer white, both are common answers under the category of color questions. Furthermore, if it's white and black, the difference between the two is greater than between white and blue, so the scoring range should be further reduced to 1-2.

Semantically dissimilar, but the answer and the standard answer mean the same under the question, then score 4-7 based on the correctness.
Examples:

(1) Question: What is lit?
Standard Answer: cake
Response: candle
Scoring: 5-7
Reason: Although cake and candle are very different semantically, in this question, they actually mean the same thing. Therefore, one should not only look at the answer but also focus on the question.

For numerical type answers, score 1-8 based on how much the number in the standard answer and the number in the response differ.
Examples:

(1) Question: How many trees are there?
Standard Answer: 4
Response: 5
Scoring: 3-5
Reason: Although 4 and 5 are different, the difference between them is not particularly large. If the standard answer is still 4, but the response becomes 1, then the scoring should be appropriately lowered to 1-3. If the standard answer is 70, and the response is 75, then it can be considered quite correct, scoring 5-7. If the standard answer is 1 and the response is 0, then score 1-2. Judge the score based on whether the numbers are relatively close to each other.

(2) Question: When did this accident happen?
Standard Answer: 1945
Response: 1940
Scoring: 5-7
Reason: The two years are quite close. But if the answer becomes the 1940s (i.e., 1941-1949), it includes 1945, and the range is not particularly large, so it is quite correct, scoring 6-8.

For answers with significantly different meanings, score based on semantic similarity without range restrictions.
Examples:

(1) Question: What color bathing suit is the woman wearing?
Standard Answer: no woman
Response: red
Scoring: 0-1
Reason: The meanings are very different.

\subsection{Manual Filter}

The following is the annotation rules provided to the three annotators during the last stage, manual filter, in the construction of our AVE dataset.

Read the question, answer, response and all augmentation results carefully, and decide whether the augmentation has changed the original meaning. Labels shall be in yes, no, unsure.

\section{Answer Templates}
In the fourth step of constructing the proposed dataset, i.e., description generation, beside collecting ChatGPT-transformed results, we augment each short response with manual written templates: (1) \textit{Answer: \{response\}.}, (2) \textit{The answer to this question is \{response\}.}, (3) \textit{As shown in the image and question, the answer is \{response\}.}, (4) \textit{The answer you are asking for is \{response\}.}, (5) \textit{As can be deduced from the image, the answer to this question is \{response\}.}, (6) \textit{The answer to your question appears to be \{response\}, as shown in the image.}.

\end{document}